\documentclass{bmvc2k}

\title{From Alignment to Synthesis: Contrastive Volumetric Grounding for Text-to-CT Generation}
\hypersetup{
  pdftitle={From Alignment to Synthesis: Contrastive Volumetric Grounding for Text-to-CT Generation},
  pdfauthor={Daniele Molino, Camillo Maria Caruso, Filippo Ruffini, Valerio Guarrasi, Paolo Soda}
}
\addauthor{Daniele Molino}{daniele.molino@unicampus.it}{1}
\addauthor{Camillo Maria Caruso}{camillomaria.caruso@unicampus.it}{1}
\addauthor{Filippo Ruffini}{filippo.ruffini@unicampus.it}{1,2}
\addauthor{Valerio Guarrasi}{valerio.guarrasi@unicamillus.org}{3}
\addauthor{Paolo Soda}{p.soda@unicampus.it}{1,2}

\addinstitution{
 Unit of Artificial Intelligence and Computer Systems,\\
 Department of Engineering,\\
 Universit\`a Campus Bio-Medico di Roma,\\
 Rome, Italy
}
\addinstitution{
 Department of Diagnostics and Intervention,\\
 Biomedical Engineering and Radiation Physics,\\
 Ume\aa{} University,\\
 Ume\aa{}, Sweden
}
\addinstitution{
 UniCamillus -- Saint Camillus International University\\
 of Health Sciences,\\
 Rome, Italy
}

\runninghead{Molino et al.}{From Alignment to Synthesis}

\usepackage{multirow, multicol, adjustbox, comment, booktabs, graphicx, amssymb}

\begin{document}

\maketitle

\begin{abstract}
Generating semantically controllable 3D CT volumes from radiology reports requires more than a rich text encoder, it requires vision-language alignment grounded in volumetric space.
Existing Text-to-CT approaches condition generation on encoders pretrained with language only or 2D vision-language objectives, providing conditioning signals that are linguistically expressive but volumetrically blind.
We argue this is a structural limitation: the quality of 3D vision-language alignment, not the richness of the text encoder, is the primary bottleneck for semantic controllability in volumetric diffusion models.
To address this, we propose a generation-oriented 3D-CLIP encoder trained with structured hard negatives that operate exclusively at the text level.
This design increases contrastive difficulty without any additional 3D memory cost, overcoming the small-batch constraints inherent to volumetric encoders. 
The resulting encoder conditions a fully end-to-end latent diffusion model that operates directly in 3D latent space, eliminating the spatial artifacts and cross-slice inconsistencies introduced by super-resolution pipelines.
Through systematic ablations, we establish a clear empirical link between grounding quality and downstream generative controllability. Evaluated on CT-RATE across 18 pathological conditions, our method achieves state-of-the-art performance on both image fidelity and factual correctness, while requiring less inference time and GPU memory than all competing methods. Code is at \url{https://github.com/danielemolino/Text2CT}.
\end{abstract}

\section{Introduction}
\label{sec1}
Text-conditioned diffusion models have demonstrated remarkable progress in 2D image generation, enabling fine-grained semantic control through natural language prompts~\cite{rombach2022high}. 
Recent works have extended this paradigm to the medical domain, showing that radiology reports can effectively guide image synthesis in 2D modalities such as chest X-rays~\cite{chambon2022roentgen,molino2025medcodi,molino2025any}.
However, transferring semantic controllability from 2D images to high-resolution 3D volumes remains largely unresolved.
Compared to planar imaging, volumetric data introduces additional structural constraints: three-dimensional scans require anatomical consistency across slices, spatial coherence along all axes, and faithful representation of subtle pathological patterns~\cite{kazerouni2023diffusion}.
Computed Tomography (CT), in particular, provides high-resolution volumetric representations widely used in clinical practice~\cite{mazonakis2016computed}, making it a challenging yet clinically representative domain for studying 3D text-conditioned medical image generation.
Yet the difficulty of this transfer is not merely structural: beyond anatomical consistency and spatial coherence, existing approaches lack a shared 3D vision-language space that is sufficiently discriminative to guide generation.
The growing interest in controllable medical image synthesis is further evidenced by the recent VLM3D challenge~\cite{challenge}, which dedicated a specific track to text-conditioned CT generation, underscoring the relevance of this problem to the broader medical vision community.

Existing Text-to-CT approaches~\cite{hamamci2024generatect,xu2024medsyn,amirrajab2025radiology} have made this task progressively more feasible, yet all share a common structural limitation: they condition generation on text encoders pretrained with language-only or 2D vision-language objectives, without any grounding in volumetric visual representations.
As a result, the conditioning signal, however rich linguistically, carries no direct information about the 3D anatomy it is meant to describe.
We argue this is a structural limitation: semantic controllability depends on \emph{volumetric contrastive grounding}, yet existing vision-language volumetric models~\cite{hamamci2024developing,blankemeier2024merlin} are constrained by the high memory footprint of 3D encoders, forcing small batch sizes where standard in-batch negatives are insufficiently discriminative.
Moreover, the reliance on image fidelity metrics such as FID~\cite{heusel2017gans} to assess these models obscures a deeper failure: a generative model can produce visually plausible CTs while systematically ignoring the conditioning report, a failure mode that only task-oriented metrics such as pathology classification can expose.

To address these limitations, we propose to improve volumetric grounding quality by augmenting contrastive training with structured \emph{hard negatives}: confusable reports that force the model to disambiguate clinically meaningful variations.
Crucially, since adding CT volumes to the batch is computationally prohibitive, our hard negatives operate exclusively at the text level, increasing contrastive signal without any additional 3D memory cost.
This design is motivated by a key property of radiology reports that makes them suited for text-level hard negative mining: clinically distinct findings can differ by as little as a single attribute, such as laterality, severity, or affected lobe, resulting in a high semantic density where minimal textual variation encodes maximal clinical difference. 
This intrinsic structure of medical language makes text-level perturbations not merely a computational convenience, but a principled and expressive source of discriminative signal.
This grounded encoder is then used to condition a fully volumetric latent diffusion model operating in 3D latent space, avoiding the artifacts introduced by super-resolution pipelines, and thus providing a controlled setting to analyze how grounding quality propagates into generative controllability.

\noindent Our contributions are as follows:
\begin{itemize}
    \item We establish an empirical link between volumetric vision-language grounding quality and downstream generative controllability, demonstrating that generation-oriented contrastive training with structured hard negatives yields more semantically controllable synthesis than alternative conditioning strategies.
    \item We introduce a generation-oriented volumetric vision-language encoder trained with structured hard negatives that improve fine-grained semantic disambiguation under the small-batch constraints imposed by 3D medical imaging.
    Beyond its role as a generative conditioning signal, this encoder achieves state-of-the-art performance on zero-shot pathology classification and volumetric retrieval tasks, demonstrating its broader utility as a 3D vision-language encoder for clinical tasks.
    \item We present a fully end-to-end Text-to-CT latent diffusion framework that operates directly in volumetric latent space, eliminating spatial artifacts and cross-slice inconsistencies introduced by super-resolution pipelines.
    \item We further demonstrate that image fidelity metrics alone are insufficient to evaluate semantic controllability in text-conditioned medical image generation, and that factual correctness, measured via pathology classification on generated volumes, provides a more sensitive and clinically meaningful proxy.
\end{itemize}

\section{Related Works}
\label{sec2}
\noindent\emph{3D Medical Image Generation.}
Generating high-resolution 3D volumes remains considerably more challenging than 2D synthesis due to increased dimensionality, memory demands, and anatomical complexity~\cite{kazerouni2023diffusion, pinaya2022brain}.
Early GAN-based approaches such as HA-GAN~\cite{sun2022hierarchical} enabled coarse CT synthesis, but suffer from training instability and limited diversity~\cite{goodfellow2020generative,saxena2021generative}.
Diffusion probabilistic models~\cite{ho2020denoising} have since emerged as a more stable alternative, with Medical Diffusion~\cite{khader2023denoising} demonstrating that diffusion combined with volumetric compression via VQ-GAN~\cite{esser2021taming} improves fidelity and scalability in 3D synthesis.
MAISI~\cite{guo2025maisi} further advanced the state of the art by enabling direct high-resolution 3D CT synthesis within a single framework, incorporating anatomy-informed conditioning through organ masks to achieve spatial control, a complementary direction to text-conditioned synthesis, which instead aims to condition generation with clinical language.

\noindent\emph{Text-Conditioned 3D CT Generation.}
GenerateCT~\cite{hamamci2024generatect} and MedSyn~\cite{xu2024medsyn} were the first to achieve text-conditioned CT synthesis, adopting cascaded pipelines where a low-resolution volume is generated from a radiology report and subsequently refined via diffusion-based super-resolution.
While these works established the feasibility of the task, the super-resolution stage introduced spatial artifacts and cross-slice inconsistencies that are incompatible with the anatomical fidelity required in clinical settings~\cite{hamamci2025better}.
Moreover, both models rely on text encoders pretrained with masked language modeling~\cite{devlin2019bert}, which capture linguistic structure but are not grounded in any visual or volumetric representation, limiting their ability to condition synthesis on fine-grained pathological cues.
BTB3D~\cite{hamamci2025better} addresses the computational bottleneck by introducing a causal convolutional encoder-decoder with wavelet-based compression and a three-stage training strategy that enables direct high-resolution volumetric synthesis without cascaded upsampling.
However, BTB3D conditions generation using a general-purpose language model~\cite{raffel2020exploring} pretrained exclusively on web-scraped text, with no exposure to medical language, leaving the quality of the conditioning signal unaddressed.
Recently, Report2CT~\cite{amirrajab2025radiology} attempted to strengthen the conditioning signal by combining BioClinicalBERT~\cite{huang2019clinicalbert}, MedEmbed~\cite{balachandran2024medembed}, and BiomedVLP CXR-BERT~\cite{boecking2022making}, respectively a clinical language model, a medical text embedding model, and a 2D vision-language model pretrained on chest X-rays.
While this strategy enriches linguistic coverage, the ensemble does not introduce explicit volumetric vision-language alignment: none of these encoders has ever been exposed to 3D CT data, and the conditioning signal therefore carries no direct information about the volumetric anatomy it is meant to guide.

Across all these works, a common assumption persists: enriching the text encoder, whether through domain adaptation or multi-encoder ensembling, is sufficient to achieve semantic controllability in volumetric synthesis.
We challenge this assumption and argue that what is missing is not richer text representations, but explicit grounding of those representations in volumetric space.

\noindent\emph{Vision-Language Grounding for Medical Imaging.}
Contrastive vision-language pretraining~\cite{radford2021learning} has shown strong results in 2D medical imaging~\cite{zhang2023biomedclip, molino2025any, zhao2023clip, you2023cxr}, however, semantic representations from planar pretraining do not transfer well to volumetric structure~\cite{kamath2023text}.
Extending contrastive alignment to CT, CT-CLIP~\cite{hamamci2024developing} trains a volumetric encoder on CT-report pairs via in-batch contrastive learning, providing a strong baseline for retrieval and zero-shot classification.
Nevertheless, standard in-batch negative sampling is particularly limiting in 3D medical setting: volumetric encoders have a high memory footprint, forcing the use of small batch sizes where random negatives are likely to be trivially dissimilar, yielding weak contrastive gradients and thus insufficiently discriminative representations~\cite{robinson2020contrastive,kalantidis2020hard}.
Merlin~\cite{blankemeier2024merlin} takes a different direction, pretraining a volumetric foundation model across heterogeneous clinical tasks, which distributes the alignment signal over multiple objectives and is therefore not optimized for the fine-grained grounding quality required to condition a generative model.
In contrast, our encoder is designed specifically with generation-oriented grounding as the primary objective, using structured hard negatives that increase contrastive difficulty at the text level, to improve fine-grained semantic disambiguation under the memory constraints inherent to 3D imaging.

\section{Methods}
\label{sec3}
\begin{figure}[ht]
    \centering
    \includegraphics[width=1\linewidth]{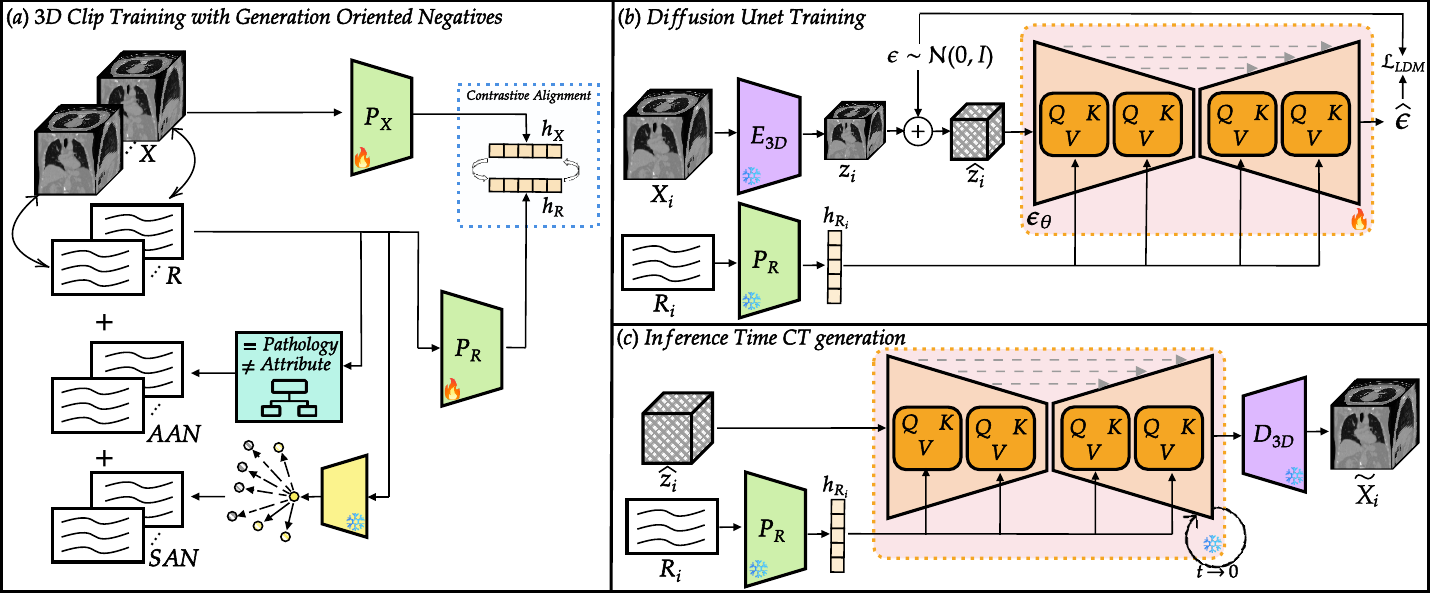}
    \caption{
        Overview of the proposed Text-to-CT generation framework: (a) \textbf{3D-CLIP Training with Generation-Oriented Negatives}: A contrastive learning setup aligns CT scans and radiology reports into a shared embedding space. To overcome the limitations of standard in-batch negatives under small batch sizes, training is augmented with two complementary miners: Attribute-Aware Negatives (AAN), which select reports sharing the same pathology but differing in clinical attributes such as laterality, severity, or affected lobe, and Semantic-Aware Negatives (SAN), which retrieve the most semantically confusable reports via nearest-neighbor search in a frozen text embedding space. (b) \textbf{Diffusion UNet Training}: A latent diffusion model is trained to denoise compressed CT representations, conditioned on textual embeddings via cross-attention. A pretrained VAE encoder compresses the volumes into latent vectors, which are noised and passed through a 3D U-Net. (c) \textbf{Inference}: A synthetic latent code is generated from noise using the textual prompt, then decoded into a high-resolution CT volume via the VAE decoder.}    
    \label{fig:model}
\end{figure}
We propose a unified generative architecture for synthesizing high-resolution 3D CT volumes from clinical descriptions, as illustrated in Figure~\ref{fig:model}.
At the core of our method is a 3D-CLIP module, shown in Figure~\ref{fig:model}.a, trained through contrastive learning to project CT volumes and radiology reports into a shared embedding space. 
To address the limitations of standard in-batch negative sampling under small batch sizes, we augment contrastive training with two complementary hard negative miners that operate exclusively at the text level, increasing contrastive difficulty without any additional 3D memory cost.
The resulting text embeddings condition a latent diffusion model that operates on a compact representation obtained via a pretrained VAE (Figure~\ref{fig:model}.b-c), eliminating the need for super-resolution stages.
We describe each component in detail below.

\subsection{Vision-Language Alignment via 3D-CLIP}
As depicted in Figure~\ref{fig:model}.a, we train a dual-encoder model comprising a 3D vision encoder $P_X(\cdot)$ and a text encoder $P_R(\cdot)$, which project input volumes $\mathbf{X}$ and reports $\mathbf{R}$ into embeddings $\mathbf{h_X}$ and $\mathbf{h_R}$, respectively.
The base alignment is trained using a symmetric InfoNCE loss~\cite{oord2018representation}.
For a batch of $N$ CT-report pairs, the image-to-text loss is defined as:
\begin{equation}
    \mathcal{L}_{\mathbf{X},\mathbf{R}} = - \frac{1}{N}\sum_{i=1}^{N}\log \frac{\exp ({\mathbf{h}_\mathbf{X_i}}^{\top} \mathbf{h}_\mathbf{R_i} / \tau)}{\sum_{j=1}^{N} \exp ({\mathbf{h}_\mathbf{X_i}}^{\top} \mathbf{h}_\mathbf{R_j} / \tau)},
\end{equation}
where $\tau$ is a learnable temperature parameter.
The symmetric text-to-image loss $\mathcal{L}_{\mathbf{R},\mathbf{X}}$ is computed analogously by treating reports as queries and CT volumes as candidates.

\begin{figure}[t]
\centering
\includegraphics[width=1\textwidth]{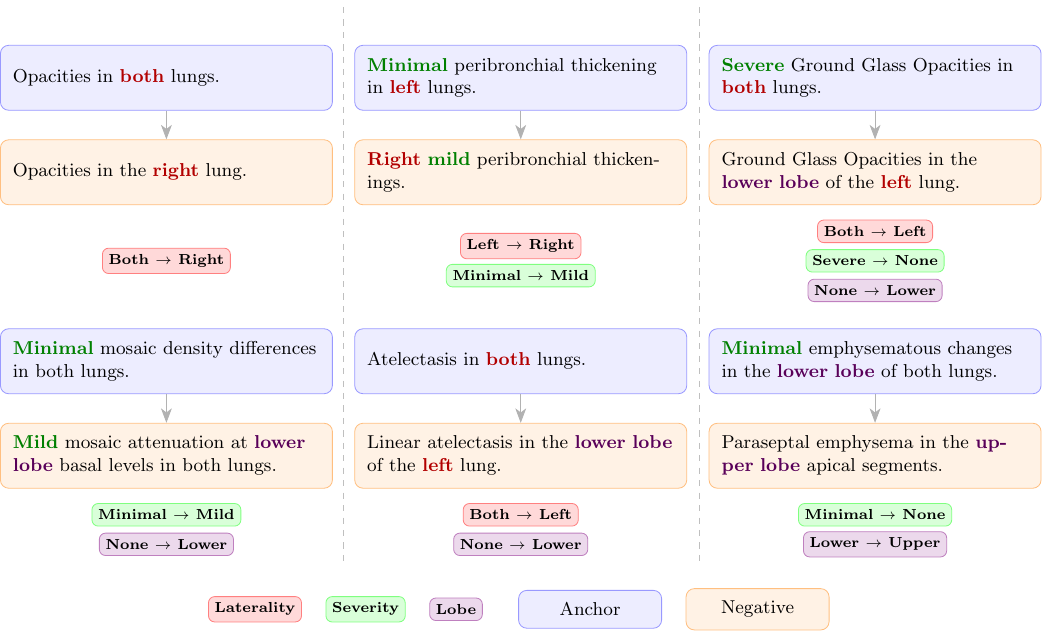}
\caption{Representative AAN pairs. Each pair shares identical pathological classes; modified attributes are color-coded: \textcolor{red!70!black}{\textbf{laterality}}, \textcolor{green!50!black}{\textbf{severity}}, \textcolor{violet!70!black}{\textbf{lobe}}.}
\label{fig:aan_pairs}
\end{figure}

\paragraph{Generation-Oriented Hard Negative Mining.}
Standard in-batch contrastive learning suffers from a critical limitation in 3D medical setting: the high memory footprint of volumetric encoders forces the use of small batch sizes, where random negatives are likely to be trivially dissimilar and thus yield weak contrastive gradients~\cite{robinson2020contrastive,kalantidis2020hard}.
To address this, we propose two hard negative miners: \textit{Attribute-Aware Negatives (AAN)} and \textit{Semantic-Aware Negatives (SAN)}, that augment each batch with additional reports without introducing any CT volume.
AAN target fine-grained clinical discrimination by selecting, for each anchor report $\mathbf{R}_i$, a set $\mathcal{A}_i$ of reports that describe a similar set of pathologies but differ in structured clinical attributes.
These negatives are \emph{clinically similar but not identical} to the anchor, forcing the encoder to separate representations that share pathological content but differ in clinically meaningful ways.
Attributes are extracted from the impression section via negation-aware lexical matching against curated sets covering three clinical dimensions: laterality (i.e., left, right, bilateral), severity (i.e., mild, moderate, severe), and lobar localization; candidates are further filtered by a label overlap constraint $|A \triangle B| \leq 3$ to ensure clinical relatedness, additional implementation details are provided in Supplementary Sec.~D.
Representative AAN pairs are illustrated in Figure~\ref{fig:aan_pairs}.

SAN target embedding-space discrimination by selecting, for each anchor $\mathbf{R}_i$, a set $\mathcal{S}_i$ of its nearest neighbors in a frozen text embedding space, filtered to exclude true positives.
These negatives are \emph{confusable reports} and pushing them away from the anchor directly improves the discriminative quality of the learned representations.
For instance, two reports such as \textit{``Millimetric nodule in the left lung''} and \textit{``Atelectasis in the left lung''} can be nearest neighbors in a clinical text embedding space with cosine similarity $0.964$, yet describe clinically distinct conditions; a generative model conditioned on such representations would receive near-identical conditioning signals for fundamentally different pathologies.
Additional SAN pairs with similarity scores and implementation details are provided in Supplementary Sec.~D.

Since hard negatives are exclusively text-level augmentations, they can only be incorporated in the image-to-text direction, where each image queries against an extended pool of report candidates.
We therefore define an augmented image-to-text loss where the denominator explicitly accounts for three sources of negative signal: in-batch reports, AAN, and SAN:
\begin{equation}
\mathcal{L}_{\mathbf{X},\mathbf{R}}^{\text{aug}} = - \frac{1}{N}\sum_{i=1}^{N} \log \frac{\exp(\mathbf{h}_\mathbf{X_i}{}^\top \mathbf{h}_{\mathbf{R_i}} / \tau)}{Z_i},
\end{equation}
where
\begin{equation}
Z_i = \sum_{j=1}^{N} \exp\!\left(\mathbf{h}_\mathbf{X_i}{}^\top \mathbf{h}_{\mathbf{R_j}} / \tau\right) + \sum_{k \in \mathcal{A}_i} \exp\!\left(\mathbf{h}_\mathbf{X_i}{}^\top \mathbf{h}_{\mathbf{R_k}} / \tau\right) + \sum_{l \in \mathcal{S}_i} \exp\!\left(\mathbf{h}_\mathbf{X_i}{}^\top \mathbf{h}_{\mathbf{R_l}} / \tau\right).
\end{equation}
The final training objective combines the augmented image-to-text loss with the standard text-to-image loss to preserve bidirectional alignment:
\begin{equation}
    \mathcal{L}_{\text{CLIP}} = \mathcal{L}_{\mathbf{X},\mathbf{R}}^{\text{aug}} + \mathcal{L}_{\mathbf{R},\mathbf{X}}.
\end{equation}
Once trained, the text encoder $P_R(\cdot)$ is frozen and used to extract conditioning embeddings for the generative model, ensuring that clinical prompts capture semantic information grounded in volumetric visual representations.

\subsection{Latent Space Compression via VAE}
To enable efficient 3D generation, we adopt a volumetric VAE~\cite{kingma2013auto} to compress CT volumes into a lower-dimensional latent space. 
The encoder $E_{3D}(\cdot)$ maps the input scan $\mathbf{X_i}$ to a latent representation $\mathbf{z_i} = E_{3D}(\mathbf{X_i})$, which is decoded back by $D_{3D}(\cdot)$ to reconstruct the original volume $\widetilde{\mathbf{X_i}} = D_{3D}(\mathbf{z_{i}})$. 
We use the 3D VAE from MAISI~\cite{guo2025maisi}, pretrained on a diverse set of CT and MRI scans using a composite objective combining voxel-wise reconstruction loss $\mathcal{L}_{\text{recon}}$, perceptual similarity $\mathcal{L}_{\text{LPIPS}}$~\cite{zhang2018unreasonable}, an adversarial term $\mathcal{L}_{\text{adv}}$~\cite{goodfellow2020generative}, and KL divergence regularization $\mathcal{L}_{\text{KL}}$ to promote smooth latent distributions~\cite{rombach2022high}.
Rather than reusing the 3D-CLIP image encoder $P_X(\cdot)$ in this stage, we adopt a dedicated pretrained VAE, as contrastive learning encourages global semantic alignment by collapsing representations of similar inputs, whereas VAE encoding preserves local spatial detail for faithful volume reconstruction, objectives that, if combined, 
would require a difficult weighting between reconstruction and generative capabilities~\cite{rombach2022high}.
The VAE is kept frozen throughout all our experiments; reconstruction quality on our test set confirms that the latent space preserves clinically relevant detail faithfully (PSNR $= 38.87$ dB, SSIM $= 0.963$), with full per-volume statistics reported in Supplementary Sec.~A.
The encoder produces latent tensors of size $\frac{H}{4} \times 
\frac{W}{4} \times \frac{D}{4}$, enabling compact but expressive volumetric representations.

\subsection{Text-Conditioned Diffusion in Latent Space}
Our generative model follows the latent diffusion formulation~\cite{rombach2022high}, operating on compressed CT representations $\mathbf{z_{i}}$ obtained via $E_{3D}(\cdot)$. 
As shown in Figure~\ref{fig:model}.b, during training, Gaussian noise is progressively added to $\mathbf{z_{i}}$ across $T$ time steps, following a predefined variance schedule $\{\beta_t\}_{t=1}^T$, with $\bar{\alpha}_t = \prod_{s=1}^{t}(1 - \beta_s)$.
The denoising network, parameterized as a 3D U-Net $\epsilon_\theta$, is trained to predict the added noise $\epsilon$ conditioned on the report embedding 
$\mathbf{h_{R_i}}$, which is injected via cross-attention~\cite{vaswani2017attention} 
at multiple resolutions. 
The training objective follows the reparameterization proposed in~\cite{ho2020denoising}, and is defined as:
\begin{equation}
\mathcal{L}_{\text{LDM}} = \mathbb{E}_{z,\varepsilon,t} 
\left[\left\|\varepsilon - \varepsilon_\theta(\hat{\mathbf{z}}_i, t, 
\mathbf{h_{R_i}})\right\|_2^2\right], \quad 
\hat{\mathbf{z}}_i = \sqrt{\bar{\alpha}_t}\mathbf{z}_i + 
\sqrt{1-\bar{\alpha}_t}\,\varepsilon, \quad \varepsilon \sim 
\mathcal{N}(0,I).
\end{equation}
Classifier-free guidance~\cite{ho2022classifier} is integrated into both training and inference to enhance the conditioning fidelity without requiring an external classifier. 
During training, the conditioning signal $\mathbf{h_{R_i}}$ is randomly omitted with a fixed probability, allowing $\epsilon_\theta$ to jointly learn conditional and unconditional generation; at inference, a guidance scale $w$ controls the strength of adherence to the conditioning report.
As shown in Figure~\ref{fig:model}.c, inference begins by sampling $\mathbf{\hat{z}_{i}}$ from a Gaussian distribution, which is iteratively denoised using the learned reverse process~\cite{song2021denoising}, guided by the report embedding $\mathbf{h_{R_i}}$. 
Finally, after $T$ denoising steps, the clean latent $\mathbf{z_{i}}$ is decoded into a full-resolution CT volume via the pretrained VAE decoder $D_{3D}$.

\section{Experimental Configuration}
\label{sec4}
\subsection{Dataset and Preprocessing}
We base our experiments on CT-RATE~\cite{hamamci2024developing}, the largest publicly available collection of paired 3D chest CT volumes and reports, spanning 18 pathologies, which is the established benchmark for text-conditioned 3D CT generation, adopted by all competing methods~\cite{hamamci2024generatect, xu2024medsyn, amirrajab2025radiology} and recognized through the VLM3D challenge~\cite{challenge} as the reference evaluation protocol for this task, following the official dataset split.
All volumes were resampled to 0.75 $\times$ 0.75 $\times$ 1.5 mm voxel spacing and cropped or padded to 512 $\times$ 512 $\times$ 128 voxels.
Hounsfield Unit values were clipped to $[-1000, +1000]$~\cite{denotter2019hounsfield, lamba2014ct} and linearly normalized to $[0, 1]$.

\subsection{Implementation Details}
\label{sec:implementation}
The vision branch is implemented as a volumetric transformer, following~\cite{hamamci2024developing}, designed to extract 3D features by jointly leveraging intra-slice and inter-slice information~\cite{kumar2024flexible, vu2020evaluation}.
The text encoder follows a masked self-attention architecture inspired by BERT~\cite{devlin2019bert}, consistent with the design adopted in CLIP~\cite{radford2021learning} and CT-CLIP~\cite{hamamci2024developing}.
For AAN, we select up to $K_A = 2$ reports per anchor; for SAN, embeddings are computed via BioClinicalBERT~\cite{huang2019clinicalbert} and the top-$K_S = 2$ neighbors are retrieved per anchor, filtering out exact matches.
BioClinicalBERT is chosen as the frozen text embedding space for SAN retrieval as, being a purely textual clinical encoder, it avoids introducing any visual bias that could circularly influence the encoder being trained.
The vision and text encoders of the 3D-CLIP were jointly trained for 100 epochs using AdamW~\cite{loshchilov2017decoupled} with learning rate $5 \times 10^{-5}$, weight decay $1 \times 10^{-4}$ and batch size equal to 4, incremented to an effective negative pool of 20 candidates per image thanks to the two miners, with negligible computational overhead as demonstrated in Supplementary Sec.~D.3.
The generative model is a 3D U-Net with cross-attention conditioning, trained for 200 epochs using AdamW~\cite{loshchilov2017decoupled} with learning rate $1 \times 10^{-4}$, batch size 8 and a polynomial learning rate decay schedule.
Inference uses DDIM sampling~\cite{song2021denoising} with 50 steps.
Classifier-free guidance was applied with a 10\% conditioning dropout probability during training; we set $w = 5.0$ across all experiments, though the effect of guidance scale on the trade-off between image fidelity and semantic controllability is analyzed in Section~\ref{sec:guidance-scale}.
All experiments were conducted on a cluster equipped with NVIDIA A100 GPUs.

\subsection{Competitors}
We compare against all methods with publicly available implementations.
For volume generation, we evaluate against GenerateCT~\cite{hamamci2024generatect}, MedSyn~\cite{xu2024medsyn}, and Report2CT~\cite{amirrajab2025radiology}.
Additionally, we benchmark our 3D-CLIP against CT-CLIP~\cite{hamamci2024developing}, Merlin~\cite{blankemeier2024merlin}, and BioMedCLIP~\cite{zhang2023biomedclip}.

\subsection{Evaluation Metrics}
\label{eval_metrics}
\noindent\emph{Image Fidelity.}
We compute FID~\cite{heusel2017gans} using two complementary approaches: \textbf{2.5D FID}, computed on axial, coronal, and sagittal slices using a RadImageNet-ResNet50~\cite{mei2022radimagenet} and \textbf{3D FID}, computed on full volumes using a 3D medical imaging backbone~\cite{chen2019med3d}.

\noindent\emph{Factual Correctness.}
We evaluate \textbf{Factual Correctness} via CT-Net~\cite{draelos2021machine}, a 3D CNN trained on CT-RATE to predict the presence of 18 pathological classes on generated volumes, quantifying how effectively each model encodes diagnostically relevant information.

\noindent\emph{Vision-Language Alignment.}
Following CT-CLIP~\cite{hamamci2024developing}, we evaluate vision-language alignment on three tasks: \textbf{Zero-Shot classification}~\cite{tiu2022expert}, where pathology presence is predicted via cosine similarity between volume embeddings and binary prompt pairs; \textbf{Volume-to-Volume retrieval}, measured by MAP@K~\cite{chen2022fast} using IoU between pathology label vectors as relevance criterion; and \textbf{Report-to-Volume retrieval}, measured by Recall@K~\cite{zhang2012query}.

\noindent\emph{Additional Analysis.}
We conduct two ablation studies to isolate the contributions of our design choices.
First, to assess the impact of volumetric vision-language grounding on generation, we replace our 3D CLIP text encoder with CT-CLIP~\cite{hamamci2024developing}, Merlin~\cite{blankemeier2024merlin}, BioMedCLIP~\cite{zhang2023biomedclip}, and BioClinicalBERT~\cite{huang2019clinicalbert} as a text-only baseline, evaluating each on FID and Factual Correctness.
Second, to quantify the contribution of each miner, we evaluate four configurations (\texttt{baseline}, \texttt{+AAN}, \texttt{+SAN}, \texttt{+AAN+SAN}) and vary $K \in \{0, 1, 2, 4, 8, 16\}$ independently for each miner, measuring the impact on downstream generation quality via FID~3D and AUC.
Additionally, we report inference time and peak GPU memory for all generative models in Table~\ref{tab:fid-efficiency}, and analyze the effect of the classifier-free guidance scale on the trade-off between image fidelity and semantic controllability in Section~\ref{sec:guidance-scale}.

\section{Results and Discussion}
\label{sec6}
This section presents an in-depth analysis of the proposed method’s performance.
Results are presented in the same order used to present the different evaluation metrics in Section~\ref{eval_metrics}.

\subsection{Image Fidelity and Computational Efficiency}
\begin{table*}[t]
\centering
\begin{adjustbox}{width=1\textwidth}
\setlength{\tabcolsep}{6pt}
\begin{tabular}{l|cccc|c|ccc}
\toprule
\multirow{2}{*}{\textbf{Method}} & \multicolumn{4}{c|}{\textbf{FID 2.5D $\downarrow$}} & \multirow{2}{*}{\textbf{FID 3D $\downarrow$}} & \textbf{Params $\downarrow$} & \textbf{Time $\downarrow$} & \textbf{VRAM $\downarrow$} \\
& \textbf{Axial} & \textbf{Coronal} & \textbf{Sagittal} & \textbf{Avg.} & & (M) & (s/vol) & (GB) \\
\midrule
GenerateCT~\cite{hamamci2024generatect} & 6.701* & 11.793* & 9.694* & 9.396* & 0.166* & \textbf{132.3} & 58.13 & 42.83 \\
MedSyn~\cite{xu2024medsyn} & 8.789* & 8.900* & 8.717* & 8.802* & 0.169* & 774.5 & 191.37 & 31.25 \\
Report2CT~\cite{amirrajab2025radiology} & \underline{1.675*} & \underline{2.694*} & \underline{2.155*} & \underline{2.175*} & 0.053* & 586.6 & \underline{46.35} & \underline{19.49} \\
Ours & \textbf{0.542} & \textbf{0.587} & \textbf{0.562} & \textbf{0.564} & \textbf{0.015} & \underline{238.6} & \textbf{24.26} & \textbf{19.43} \\
\bottomrule
\end{tabular}
\end{adjustbox}
\caption{FID scores and computational efficiency across Text-to-CT methods. FID is computed using 2.5D slice-based and 3D volumetric feature extractors; lower values indicate greater similarity to real CT distributions. Computational metrics include number of parameters (M), mean inference time per generated volume (s), and peak GPU memory (GB). FID results are averaged over 10 runs with different random seeds. Best results are in \textbf{bold}, second-best \underline{underlined}. * indicates statistical significance vs.\ our method ($p < 0.05$, Wilcoxon signed-rank test across 10 runs).}
\label{tab:fid-efficiency}
\end{table*}

\begin{figure}[h]
    \centering
    \includegraphics[width=1\textwidth]{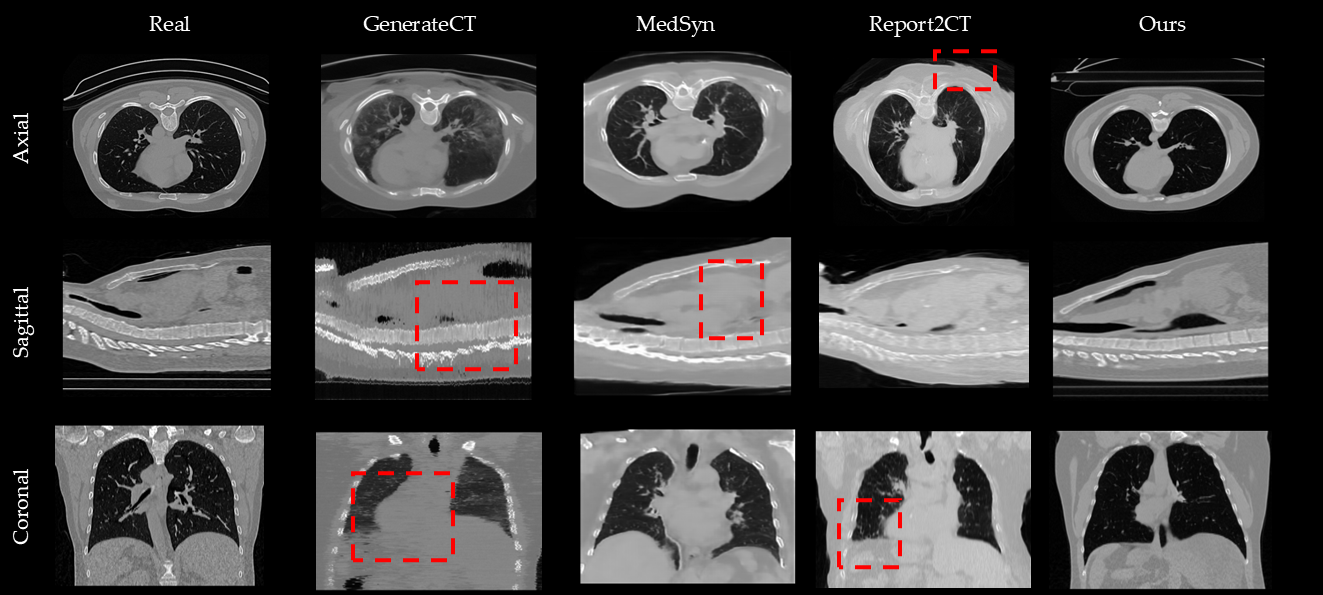}
    \caption{Qualitative comparison of real and synthetic CT volumes generated 
    by competing methods using the same textual prompt. Rows show axial, 
    sagittal, and coronal slices. Red boxes highlight 
    regions affected by visual artifacts introduced by super-resolution stages.}
    \label{fig:qual}
\end{figure}

Table~\ref{tab:fid-efficiency} jointly reports generation quality and inference efficiency across all compared methods.
In terms of image fidelity, our model achieves the lowest FID across all views and backbones, indicating a closer match to the real CT distribution.
GenerateCT shows relatively good performance along the axial plane, but degrades significantly in the coronal and sagittal directions, reflecting the spatial incoherence introduced by its 2D super-resolution.
MedSyn applies 3D super-resolution, yielding uniform scores, but still introduces grid-like artifacts, as illustrated in Figure~\ref{fig:qual}.
Report2CT achieves competitive fidelity, benefiting from its richer text encoder ensemble, yet still falls short of our approach across all metrics.
In the 3D FID evaluation, which captures volumetric realism of entire scans, our method outperforms all competitors by a large margin, confirming that operating directly in volumetric latent space avoids the artifacts introduced by cascaded pipelines.
In terms of computational efficiency, our model achieves the fastest inference and the lowest peak GPU memory, despite operating at full resolution without any super-resolution stage.
Figure~\ref{fig:qual} presents a qualitative comparison of CT volumes generated by each method using the same textual prompt. 
Additional qualitative comparisons are provided in Supplementary Sec.~E.

\subsection{Factual Correctness Evaluation}
\begin{table}[t]
\centering
\begin{adjustbox}{width=1\textwidth}
\begin{tabular}{l|cccc}
\toprule
\textbf{Model} & \textbf{AUC (Macro)} & \textbf{AUC (Weighted)} & \textbf{Precision (Macro)} & \textbf{Precision (Weighted)} \\
\midrule
GenerateCT~\cite{hamamci2024generatect} & 0.541* & 0.529* & 0.285* & 0.359* \\
MedSyn~\cite{xu2024medsyn} & 0.520* & 0.517* & 0.282* & 0.361* \\
Report2CT~\cite{amirrajab2025radiology} & \underline{0.720*} & \underline{0.705*} & \underline{0.436*} & \underline{0.508*} \\
Ours & \textbf{0.787} & \textbf{0.776} & \textbf{0.535} & \textbf{0.605} \\
\bottomrule
\end{tabular}
\end{adjustbox}
\caption{Evaluation of factual correctness using CT-Net~\cite{draelos2021machine}. Table reports AUC and Precision across 18 disease classes on synthetic test sets generated by each method. For reference, CT-Net achieves AUC (Macro) = 0.821 on real test data. Best results are in \textbf{bold}, second-best \underline{underlined}. * indicates statistical significance vs.\ our method ($p < 0.05$, Wilcoxon signed-rank test on per-volume CT-Net scores).}
\label{tab:ctnet-results}
\end{table}
To assess the factual correctness of the generated CT volumes, we evaluated their diagnostic informativeness using CT-Net~\cite{draelos2021machine}. 
Table~\ref{tab:ctnet-results} reports AUC and Precision as both macro and weighted averages over 18 disease classes.
Our model achieves the highest AUC and Precision across both aggregation strategies, demonstrating its ability to produce diagnostically meaningful CT volumes that faithfully encode the pathological content described in the conditioning reports.
Report2CT achieves the second-best results, reflecting the benefit of its domain-adapted text encoder ensemble, yet a consistent gap remains across all metrics, suggesting that linguistic coverage alone is insufficient without explicit volumetric vision-language grounding.
GenerateCT and MedSyn show the weakest performance, likely due to their misaligned text encoders and the structural artifacts introduced by their super-resolution pipelines.
A complete per-class breakdown is provided in Supplementary Sec.~B.
We further evaluate the utility of synthetic volumes for downstream classifier training under real-only, synthetic-only, and combined regimes; results are reported in Supplementary Sec.~C.

\subsection{Vision-Language Alignment}
\label{sec:vla}
\begin{table*}[t]
\centering
\begin{adjustbox}{width=1\textwidth}
\setlength{\tabcolsep}{6pt}
\begin{tabular}{l|cc|cccc|cccc}
\toprule
\multirow{2}{*}{\textbf{Methods}} 
& \multicolumn{2}{c|}{\textbf{Zero-Shot}} 
& \multicolumn{4}{c|}{\textbf{Volume-to-Volume (MAP@K)}} 
& \multicolumn{4}{c}{\textbf{Report-to-Volume (Recall@K)}} \\
& \textbf{AUC $\uparrow$} & \textbf{Pre. $\uparrow$} 
& \textbf{MAP@1 $\uparrow$} & \textbf{MAP@5 $\uparrow$} & \textbf{MAP@10 $\uparrow$} & \textbf{MAP@50 $\uparrow$} 
& \textbf{Re@5 $\uparrow$} & \textbf{Re@10 $\uparrow$} & \textbf{Re@50 $\uparrow$} & \textbf{Re@100 $\uparrow$} \\
\midrule
Random & 0.500 & 0.181 & 0.185 & 0.162 & 0.155 & 0.146 & 0.002 & 0.005 & 0.025 & 0.049 \\
\midrule
BioMedCLIP$^\dagger$~\cite{zhang2023biomedclip} & 0.544 & 0.256 & 0.360 & 0.253 & 0.226 & 0.186 & 0.004 & 0.008 & 0.035 & 0.066 \\
Merlin~\cite{blankemeier2024merlin} & 0.623 & 0.290 & 0.474 & 0.302 & 0.241 & 0.222 & 0.012 & 0.017 & 0.022 & 0.177 \\
CT-CLIP~\cite{hamamci2024developing}            & \underline{0.625} & \underline{0.313} & \underline{0.837} & \underline{0.450} & \underline{0.302} & \underline{0.244} & \underline{0.029} & \underline{0.051} & \underline{0.183} & \underline{0.288} \\
Ours & \textbf{0.676} & \textbf{0.353} & \textbf{0.866} & \textbf{0.498} & \textbf{0.342} & \textbf{0.289} & \textbf{0.041} & \textbf{0.072} & \textbf{0.229} & \textbf{0.368} \\
\bottomrule
\end{tabular}
\end{adjustbox}
\caption{Comparison across zero-shot classification (AUC and Precision), volume-to-volume retrieval (MAP@K), and report-to-volume retrieval (Recall@K). Best results are in \textbf{bold}, second-best \underline{underlined}. $^\dagger$BioMedCLIP is applied to CT volumes by encoding non-overlapping 16-slice chunks and averaging the resulting embeddings.}
\label{tab:clip-eval}
\end{table*}

Table~\ref{tab:clip-eval} reports the comparative evaluation of our 3D-CLIP model across three vision-language alignment tasks.
Our encoder is trained on the same dataset as CT-CLIP using the same contrastive objective, differing solely in the addition of AAN and SAN under small-batch constraints, making it the most direct point of comparison for isolating the contribution of our generation-oriented hard negative mining strategy.
In the zero-shot classification setting, our model achieves the best results.
CT-CLIP represents the closest competitor, while Merlin lags behind on both metrics despite its broad multi-task pretraining, which may suggest that distributing the alignment signal over heterogeneous objectives is suboptimal for fine-grained pathology recognition.
BioMedCLIP, applied by encoding 16-slice chunks independently and averaging embeddings, shows the weakest zero-shot performance, confirming that 2D representations do not generalize to volumetric anatomy.
In the volume-to-volume retrieval task, our model outperforms all baselines across all MAP@K values.
The improvement is most pronounced at MAP@1, indicating that our embedding space more consistently ranks the most semantically similar volume at the top of the retrieval list.
This is the retrieval regime most directly affected by AAN: when two volumes share a primary pathology but differ in structured attributes, our encoder has been explicitly trained to assign higher similarity to the more clinically aligned candidate.
In the report-to-volume retrieval setting, our model yields higher Recall@K scores across all values of $K$, with the most pronounced gains at larger $K$ (Re@100: 0.368 vs.\ 0.288 for CT-CLIP).
This is precisely the regime targeted by SAN, which augments training with the nearest neighbors in text embedding space and increases separation between the most confusable report pairs.
These results anticipate a broader trend that is examined systematically in Section~\ref{sec:ablation-encoder}: encoders that achieve stronger vision-language alignment yield more semantically controllable generation, suggesting that grounding quality propagates directly into downstream generative performance.

\subsection{Ablation on Contrastive Vision-Language Pretraining}
\label{sec:ablation-encoder}
\begin{table}[t]
\small
\centering
\begin{adjustbox}{width=0.8\textwidth}
\begin{tabular}{l|cc|cc}
\toprule
\textbf{Text Encoder} & \textbf{FID 2.5D Avg. $\downarrow$} & \textbf{FID 3D $\downarrow$} & \textbf{AUC $\uparrow$} & \textbf{Precision $\uparrow$} \\
\midrule
BioClinicalBERT~\cite{huang2019clinicalbert} & 0.731 & 0.048 & 0.704 & 0.393 \\
BioMedCLIP~\cite{zhang2023biomedclip}        & 0.695 & 0.047 & 0.666 & 0.380 \\
Merlin~\cite{blankemeier2024merlin}          & 0.651 & 0.028 & 0.722 & 0.443 \\
CT-CLIP~\cite{hamamci2024developing}         & \underline{0.632} & \underline{0.026} & \underline{0.753} & \underline{0.471} \\
Ours (3D-CLIP) & \textbf{0.564} & \textbf{0.015} & \textbf{0.787} & \textbf{0.535} \\
\bottomrule
\end{tabular}
\end{adjustbox}
\caption{Ablation on text encoder conditioning. All variants use the same generative model and differ only in the text encoder used for conditioning. Best results are in \textbf{bold}, second-best \underline{underlined}.}
\label{tab:ablation}
\end{table}
To quantify the link between vision-language alignment and generative controllability suggested by the retrieval results in Section~\ref{sec:vla}, we replace our 3D-CLIP text encoder with four alternative conditioning strategies of increasing alignment quality, keeping the generative model fixed across all variants.
This controlled design is deliberate: by isolating the conditioning component while holding the generative backbone constant, FID and classification metrics become complementary probes of two distinct properties. 
FID captures distributional realism, which is dominated by the generative backbone. 
AUC and Precision, by contrast, directly measures whether the generated volume encodes the specific pathological content described in the conditioning report, a property that is entirely determined by the quality of the conditioning signal. 
Table~\ref{tab:ablation} reveals a consistent trend: as the degree of volumetric vision-language grounding increases from BioClinicalBERT to our 3D-CLIP, both FID and classification performance improve monotonically with one exception.
BioMedCLIP performs slightly worse than BioClinicalBERT despite being a vision-language model, which we hypothesize is due to the nature of its pretraining: BioMedCLIP is aligned to a heterogeneous 2D biomedical image space that includes microscopy, pathology, and X-rays, introducing a visual bias that is misaligned with CT volumetric anatomy.
As a result, its text representations carry cross-modal information that is not only unhelpful but potentially misleading for CT conditioning, whereas BioClinicalBERT, as a purely textual clinical encoder, provides a cleaner and more domain-consistent signal.
Importantly, the performance gap across encoders is more pronounced in classification than in FID, consistent with the design rationale above: a generative model can learn to replicate the statistical distribution of real CT volumes even under weak conditioning, but cannot generate pathological content that the conditioning signal fails to discriminate.
This asymmetry further supports our argument that FID alone is an insufficient evaluation criterion in the medical domain, and that factual correctness provides a more sensitive measure of semantic controllability.

\subsection{Ablation on Hard Negative Mining}
\label{sec:ablation-hn}
\begin{figure*}[t]
    \centering
    \includegraphics[width=0.9\textwidth]{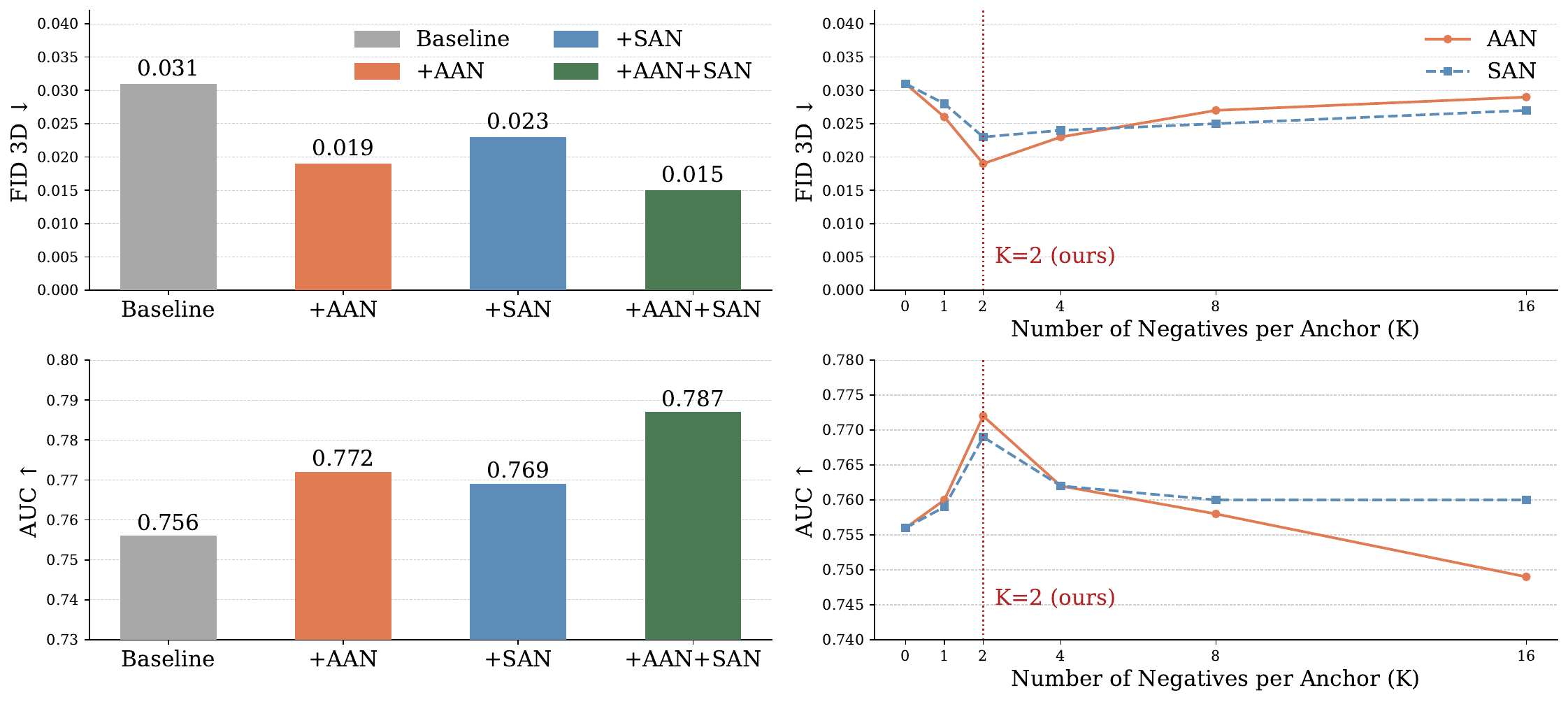}
    \caption{
        \textbf{(a) Effect of Negative Miners.}
        Adding AAN and SAN individually improves both FID 3D and AUC over the baseline,
        with the combination yielding the best performance.
        \textbf{(b) Sensitivity to Number of Negatives.}
        Each curve shows the effect of varying $K$ for one miner while the other is disabled ($K=0$).
        AAN performance peaks at $K=2$ and degrades more sharply at higher values, while SAN remains more robust to increases in $K$. The red line marks $K=2$, the value used in our final model.
    }
    \label{fig:ablation-hn}
\end{figure*}
Figure~\ref{fig:ablation-hn} reports the results of two complementary ablations on the negative mining strategy.
Panel~(a) evaluates the contribution of each miner in isolation and in combination.
Both AAN and SAN individually improve over the baseline, confirming that each miner provides a distinct and complementary source of signal that propagates into generative controllability.
The full combination achieves the best performance on both metrics, indicating that attribute-level and embedding-space discrimination are not redundant but reinforce each other.
Panel~(b) analyzes the sensitivity of each miner to the number of negatives per anchor $K$, with the other miner disabled.
For AAN, performance peaks at $K=2$ and degrades at higher values: at large $K$, it becomes increasingly difficult to find candidates that are both clinically similar and meaningfully different in structured attributes, raising the risk of including negatives that carry ambiguous or conflicting signal.
This degradation is more pronounced on AUC, where AAN drops below the baseline at $K=16$, suggesting that noisy attribute-level negatives can actively mislead the encoder when selected indiscriminately.
SAN is more robust across $K$, as nearest-neighbor ranking provides a natural quality filter that prevents degradation at higher counts.
Together, these results suggest that in the low-batch-size regime typical of 3D medical imaging, the quality of negative selection matters more than quantity, and that a small, carefully selected set of negatives is sufficient to improve downstream generative performance.
Additionally, the computational cost of this augmentation strategy remains negligible across all values of $K$, as reported in Supplementary Sec.~D.3.

\subsection{Preliminary Reader Study}
\label{sec:reader-study}
\begin{table}[t]
\centering
\begin{tabular}{l|cc}
\toprule
\textbf{Method} & \textbf{Anat. Realism $\uparrow$} & \textbf{Report Consistency $\uparrow$} \\
\midrule
Report2CT~\cite{amirrajab2025radiology}  & $3.67 \pm 0.73^*$ & $3.18 \pm 0.68^*$ \\
BioClinicalBERT (abl.)                   & $3.88 \pm 0.65^*$ & $3.04 \pm 0.72^*$ \\
Ours                                     & $\mathbf{4.08 \pm 0.63}$ & $\mathbf{3.84 \pm 0.61}$ \\
\bottomrule
\end{tabular}
\caption{Preliminary reader study results (mean $\pm$ std, $n=25$ volumes per method). A board-certified radiologist (7 years experience) rated each volume on Anatomical Realism and Report Consistency on a 1--5 Likert scale. * indicates statistical significance vs.\ our method ($p < 0.05$, Mann-Whitney U test).}
\label{tab:reader-study}
\end{table}
To complement the quantitative evaluation, we conducted a preliminary blind reader study with a board-certified radiologist.
The radiologist rated volumes generated by Ours, 
Report2CT~\cite{amirrajab2025radiology}, and the BioClinicalBERT ablation from the same set of 25 randomly sampled conditioning reports, on two criteria: Anatomical Realism and Report Consistency, each scored 
on a 1--5 Likert scale.
Table~\ref{tab:reader-study} reports the results.
Anatomical Realism scores are comparable across methods, suggesting that all three approaches produce visually plausible volumes.
Report Consistency, however, shows a more pronounced improvement for our method compared to both Report2CT and the BioClinicalBERT ablation, confirming that volumetric grounding quality drives semantic controllability in a way that is perceptible to a clinical expert.
While the scale of this study remains preliminary, these results are coherent with the quantitative findings in Table~\ref{tab:ctnet-results}.

\subsection{Effect of Guidance Scale on Generation and Clinical Utility}
\label{sec:guidance-scale}
\begin{figure}[t]
    \centering
    \includegraphics[width=0.8\linewidth]{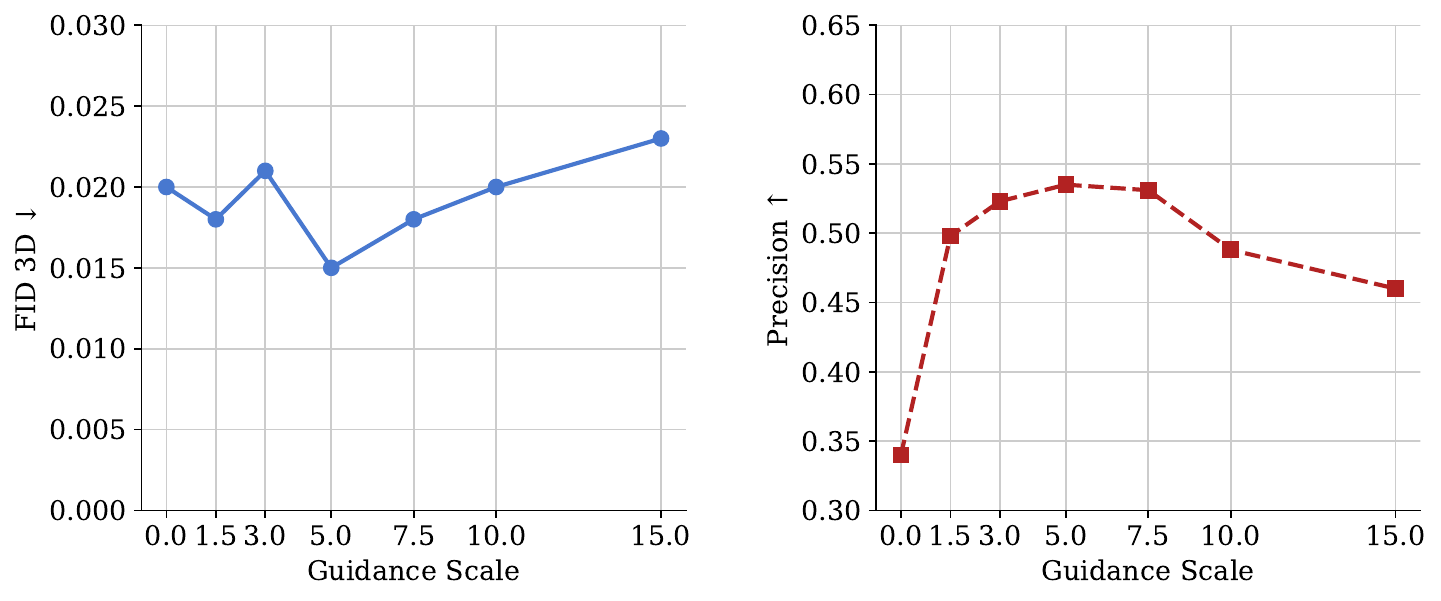}
    \caption{Effect of guidance scale on FID 3D and Precision. While FID remains low and relatively stable across scales, precision peaks at a moderate guidance value, highlighting the importance of semantic controllability beyond visual similarity.}
    \label{fig:guidance_scale}
\end{figure}
Figure~\ref{fig:guidance_scale} reports FID 3D and Precision as the classifier-free guidance scale $w$ varies from 0 to 15.
FID remains low across all settings, suggesting that visual fidelity is largely insensitive to the guidance strength.
Precision, by contrast, varies and peaks at $w = 5.0$, which we adopt as the default value in all experiments.
This result reinforces the observation made in Section~\ref{sec:ablation-encoder}, FID is an insufficient criterion for evaluating text-conditioned medical image generation: a model can maintain low FID while its semantic controllability, as measured by Precision, varies substantially, a failure mode that only task-oriented metrics expose.

\section{Conclusion}
\label{sec:conclusion}
In this work, we presented a framework for text-conditioned generation of high-resolution 3D CT volumes that places volumetric contrastive grounding at the center of the design.
Our central argument, that semantic controllability in volumetric diffusion models depends critically on the quality of vision-language alignment, is supported empirically across all evaluation axes: image fidelity, factual correctness and vision-language alignment.
The generation-oriented 3D-CLIP encoder, trained with Attribute and Semantic-Aware Negatives under the small-batch constraints inherent to 3D medical imaging, outperforms alternative conditioning strategies, with the most pronounced gains observed on metrics that directly probe semantic alignment.
The ablation studies further confirm that structured hard negatives provide complementary improvements, and that a small, carefully selected set of negatives is more effective than naive scaling of their quantity.
Beyond its role as a conditioning signal, the encoder achieves state-of-the-art performance on zero-shot pathology classification and volumetric retrieval against dedicated vision-language baselines, suggesting its utility extends to clinical tasks that do not involve generation.
A preliminary reader study further confirms that the improvement is perceptible to a clinical expert, providing initial evidence of clinical relevance beyond automated metrics.
However, this work has some limitations.
Our evaluation is conducted exclusively on CT-RATE, which limits the generalizability of our conclusions to other districts and protocols.
Additionally, hard negatives in our framework operate exclusively at the text level, a deliberate design choice motivated by the memory constraints of 3D encoders, but that precludes the use of volumetric hard negatives that could further strengthen contrastive signal.
Future work could extend this framework to other anatomical regions as paired text-volume datasets become available, incorporate volumetric hard negatives and expand the preliminary reader study to a larger-scale formal evaluation involving multiple radiologists and a broader range of pathological presentations.
\bibliography{main}

@String(NIPS= {Adv. Neural Inform. Process. Syst.})

@String(NIPS  = {NeurIPS})

@article{chambon2022roentgen,
  title={{A vision--language foundation model for the generation of realistic chest x-ray images}},
  author={Bluethgen, Christian and Chambon, Pierre and Delbrouck, Jean-Benoit and van der Sluijs, Rogier and Po{\l}acin, Ma{\l}gorzata and Zambrano Chaves, Juan Manuel and Abraham, Tanishq Mathew and Purohit, Shivanshu and Langlotz, Curtis P and Chaudhari, Akshay S},
  journal={Nature Biomedical Engineering},
  pages={1--13},
  year={2024},
  publisher={Nature Publishing Group UK London}
}

@article{molino2025medcodi,
  title={{XGeM: A multi-prompt foundation model for multimodal medical data generation}},
  author={Molino, Daniele and Di Feola, Francesco and Faiella, Eliodoro and Fazzini, Deborah and Santucci, Domiziana and Shen, Linlin and Guarrasi, Valerio and Soda, Paolo},
  journal={Computerized Medical Imaging and Graphics},
  pages={102718},
  year={2026},
  publisher={Elsevier}
}

@article{mazonakis2016computed,
  title={{Computed tomography: What and how does it measure?}},
  author={Mazonakis, Michalis and Damilakis, John},
  journal={European journal of radiology},
  volume={85},
  number={8},
  pages={1499--1504},
  year={2016},
  publisher={Elsevier}
}

@article{kazerouni2023diffusion,
  title={{Diffusion models in medical imaging: A comprehensive survey}},
  author={Kazerouni, Amirhossein and Aghdam, Ehsan Khodapanah and Heidari, Moein and Azad, Reza and Fayyaz, Mohsen and Hacihaliloglu, Ilker and Merhof, Dorit},
  journal={Medical image analysis},
  volume={88},
  pages={102846},
  year={2023},
  publisher={Elsevier}
}

@inproceedings{rombach2022high,
  title={{High-resolution image synthesis with latent diffusion models}},
  author={Rombach, Robin and Blattmann, Andreas and Lorenz, Dominik and Esser, Patrick and Ommer, Bj{\"o}rn},
  booktitle={Proceedings of the IEEE/CVF conference on computer vision and pattern recognition},
  pages={10684--10695},
  year={2022}
}

@article{sun2022hierarchical,
  title={{Hierarchical amortized GAN for 3D high resolution medical image synthesis}},
  author={Sun, Li and Chen, Junxiang and Xu, Yanwu and Gong, Mingming and Yu, Ke and Batmanghelich, Kayhan},
  journal={IEEE journal of biomedical and health informatics},
  volume={26},
  number={8},
  pages={3966--3975},
  year={2022},
  publisher={IEEE}
}

@article{khader2023denoising,
  title={{Denoising diffusion probabilistic models for 3D medical image generation}},
  author={Khader, Firas and M{\"u}ller-Franzes, Gustav and Tayebi Arasteh, Soroosh and Han, Tianyu and Haarburger, Christoph and Schulze-Hagen, Maximilian and Schad, Philipp and Engelhardt, Sandy and Bae{\ss}ler, Bettina and Foersch, Sebastian and others},
  journal={Scientific Reports},
  volume={13},
  number={1},
  pages={7303},
  year={2023},
  publisher={Nature Publishing Group UK London}
}

@article{xu2024medsyn,
  title={{Medsyn: Text-guided anatomy-aware synthesis of high-fidelity 3d ct images}},
  author={Xu, Yanwu and Sun, Li and Peng, Wei and Jia, Shuyue and Morrison, Katelyn and Perer, Adam and Zandifar, Afrooz and Visweswaran, Shyam and Eslami, Motahhare and Batmanghelich, Kayhan},
  journal={IEEE Transactions on Medical Imaging},
  year={2024},
  publisher={IEEE}
}

@inproceedings{hamamci2024generatect,
  title={{Generatect: Text-conditional generation of 3d chest ct volumes}},
  author={Hamamci, Ibrahim Ethem and Er, Sezgin and Sekuboyina, Anjany and Simsar, Enis and Tezcan, Alperen and Simsek, Ayse Gulnihan and Esirgun, Sevval Nil and Almas, Furkan and Do{\u{g}}an, Irem and Dasdelen, Muhammed Furkan and others},
  booktitle={European Conference on Computer Vision},
  pages={126--143},
  year={2024},
  organization={Springer}
}

@inproceedings{guo2025maisi,
  title={{Maisi: Medical ai for synthetic imaging}},
  author={Guo, Pengfei and Zhao, Can and Yang, Dong and Xu, Ziyue and Nath, Vishwesh and Tang, Yucheng and Simon, Benjamin and Belue, Mason and Harmon, Stephanie and Turkbey, Baris and others},
  booktitle={2025 IEEE/CVF Winter Conference on Applications of Computer Vision (WACV)},
  pages={4430--4441},
  year={2025},
  organization={IEEE}
}

@article{goodfellow2020generative,
  title={{Generative adversarial networks}},
  author={Goodfellow, Ian and Pouget-Abadie, Jean and Mirza, Mehdi and Xu, Bing and Warde-Farley, David and Ozair, Sherjil and Courville, Aaron and Bengio, Yoshua},
  journal={Communications of the ACM},
  volume={63},
  number={11},
  pages={139--144},
  year={2020},
  publisher={ACM New York, NY, USA}
}

@inproceedings{ho2020denoising,
author = {Ho, Jonathan and Jain, Ajay and Abbeel, Pieter},
title = {{Denoising diffusion probabilistic models}},
year = {2020},
publisher = {Curran Associates Inc.},
booktitle = {Proceedings of the 34th International Conference on Neural Information Processing Systems},
articleno = {574},
numpages = {12},
location = {Vancouver, BC, Canada},
series = {NIPS '20}
}

@inproceedings{esser2021taming,
  title={{Taming transformers for high-resolution image synthesis}},
  author={Esser, Patrick and Rombach, Robin and Ommer, Bjorn},
  booktitle={Proceedings of the IEEE/CVF conference on computer vision and pattern recognition},
  pages={12873--12883},
  year={2021}
}

@article{vaswani2017attention,
  title={{Attention is all you need}},
  author={Vaswani, Ashish and Shazeer, Noam and Parmar, Niki and Uszkoreit, Jakob and Jones, Llion and Gomez, Aidan N and Kaiser, {\L}ukasz and Polosukhin, Illia},
  journal={Advances in neural information processing systems},
  volume={30},
  year={2017}
}

@article{raffel2020exploring,
  title={{Exploring the limits of transfer learning with a unified text-to-text transformer}},
  author={Raffel, Colin and Shazeer, Noam and Roberts, Adam and Lee, Katherine and Narang, Sharan and Matena, Michael and Zhou, Yanqi and Li, Wei and Liu, Peter J},
  journal={Journal of machine learning research},
  volume={21},
  number={140},
  pages={1--67},
  year={2020}
}

@inproceedings{boecking2022making,
  title={{Making the most of text semantics to improve biomedical vision--language processing}},
  author={Boecking, Benedikt and Usuyama, Naoto and Bannur, Shruthi and Castro, Daniel C and Schwaighofer, Anton and Hyland, Stephanie and Wetscherek, Maria and Naumann, Tristan and Nori, Aditya and Alvarez-Valle, Javier and others},
  booktitle={European conference on computer vision},
  pages={1--21},
  year={2022},
  organization={Springer}
}

@inproceedings{devlin2019bert,
  title={{Bert: Pre-training of deep bidirectional transformers for language understanding}},
  author={Devlin, Jacob and Chang, Ming-Wei and Lee, Kenton and Toutanova, Kristina},
  booktitle={Proceedings of the 2019 conference of the North American chapter of the association for computational linguistics: human language technologies, volume 1 (long and short papers)},
  pages={4171--4186},
  year={2019}
}

@inproceedings{radford2021learning,
  title={{Learning transferable visual models from natural language supervision}},
  author={Radford, Alec and Kim, Jong Wook and Hallacy, Chris and Ramesh, Aditya and Goh, Gabriel and Agarwal, Sandhini and Sastry, Girish and Askell, Amanda and Mishkin, Pamela and Clark, Jack and others},
  booktitle={International conference on machine learning},
  pages={8748--8763},
  year={2021},
  organization={PMLR}
}

@article{hamamci2024developing,
  title={{Developing generalist foundation models from a multimodal dataset for 3d computed tomography}},
  author={Hamamci, Ibrahim Ethem and Er, Sezgin and Wang, Chenyu and Almas, Furkan and Simsek, Ayse Gulnihan and Esirgun, Sevval Nil and Doga, Irem and Durugol, Omer Faruk and Dai, Weicheng and Xu, Murong and others},
  journal={arXiv preprint arXiv:2403.17834},
  year={2024}
}

@book{denotter2019hounsfield,
	Title = {Hounsfield Unit},
	Author = {DenOtter, Tami D. and Schubert, Johanna},
	Publisher = {StatPearls Publishing, Treasure Island (FL)},
	Year = {2025},
	URL = {http://europepmc.org/books/NBK547721},
}

@article{lamba2014ct,
  title={{CT Hounsfield numbers of soft tissues on unenhanced abdominal CT scans: variability between two different manufacturers’ MDCT scanners}},
  author={Lamba, Ramit and McGahan, John P and Corwin, Michael T and Li, Chin-Shang and Tran, Tien and Seibert, J Anthony and Boone, John M},
  journal={American Journal of Roentgenology},
  volume={203},
  number={5},
  pages={1013--1020},
  year={2014},
  publisher={American Roentgen Ray Society}
}

@inproceedings{you2023cxr,
  title={{Cxr-clip: Toward large scale chest x-ray language-image pre-training}},
  author={You, Kihyun and Gu, Jawook and Ham, Jiyeon and Park, Beomhee and Kim, Jiho and Hong, Eun K and Baek, Woonhyuk and Roh, Byungseok},
  booktitle={International Conference on Medical Image Computing and Computer-Assisted Intervention},
  pages={101--111},
  year={2023},
  organization={Springer}
}

@article{zhao2023clip,
  title={{Clip in medical imaging: A comprehensive survey}},
  author={Zhao, Zihao and Liu, Yuxiao and Wu, Han and Wang, Mei and Li, Yonghao and Wang, Sheng and Teng, Lin and Liu, Disheng and Cui, Zhiming and Wang, Qian and others},
  journal={arXiv preprint arXiv:2312.07353},
  year={2023}
}

@article{kumar2024flexible,
  title={{A flexible 2.5 D medical image segmentation approach with in-slice and cross-slice attention}},
  author={Kumar, Amarjeet and Jiang, Hongxu and Imran, Muhammad and Valdes, Cyndi and Leon, Gabriela and Kang, Dahyun and Nataraj, Parvathi and Zhou, Yuyin and Weiss, Michael D and Shao, Wei},
  journal={Computers in Biology and Medicine},
  volume={182},
  pages={109173},
  year={2024},
  publisher={Elsevier}
}

@article{vu2020evaluation,
  title={{Evaluation of multislice inputs to convolutional neural networks for medical image segmentation}},
  author={Vu, Minh H and Grimbergen, Guus and Nyholm, Tufve and L{\"o}fstedt, Tommy},
  journal={Medical Physics},
  volume={47},
  number={12},
  pages={6216--6231},
  year={2020},
  publisher={Wiley Online Library}
}

@article{oord2018representation,
  title={{Representation learning with contrastive predictive coding}},
  author={Oord, Aaron van den and Li, Yazhe and Vinyals, Oriol},
  journal={arXiv preprint arXiv:1807.03748},
  year={2018}
}

@article{kingma2013auto,
  title={{Auto-encoding variational Bayes}},
  author={Kingma, Diederik P and Welling, Max},
  journal={arXiv preprint arXiv:1312.6114},
  year={2013}
}

@inproceedings{pinaya2022brain,
  title={{Brain imaging generation with latent diffusion models}},
  author={Pinaya, Walter HL and Tudosiu, Petru-Daniel and Dafflon, Jessica and Da Costa, Pedro F and Fernandez, Virginia and Nachev, Parashkev and Ourselin, Sebastien and Cardoso, M Jorge},
  booktitle={MICCAI Workshop on Deep Generative Models},
  pages={117--126},
  year={2022},
  organization={Springer}
}

@inproceedings{zhang2018unreasonable,
  title={{The unreasonable effectiveness of deep features as a perceptual metric}},
  author={Zhang, Richard and Isola, Phillip and Efros, Alexei A and Shechtman, Eli and Wang, Oliver},
  booktitle={Proceedings of the IEEE conference on computer vision and pattern recognition},
  pages={586--595},
  year={2018}
}

@article{ho2022classifier,
  title={{Classifier-Free Diffusion Guidance}},
  author={Ho, Jonathan and Salimans, Tim},
  journal={arXiv preprint arXiv:2207.12598},
  year={2022}
}

@article{loshchilov2017decoupled,
  title={{Decoupled weight decay regularization}},
  author={Loshchilov, Ilya and Hutter, Frank},
  journal={arXiv preprint arXiv:1711.05101},
  year={2017}
}

@article{heusel2017gans,
  title={{Gans trained by a two time-scale update rule converge to a local nash equilibrium}},
  author={Heusel, Martin and Ramsauer, Hubert and Unterthiner, Thomas and Nessler, Bernhard and Hochreiter, Sepp},
  journal={Advances in neural information processing systems},
  volume={30},
  year={2017}
}

@article{draelos2021machine,
  title={{Machine-learning-based multiple abnormality prediction with large-scale chest computed tomography volumes}},
  author={Draelos, Rachel Lea and Dov, David and Mazurowski, Maciej A and Lo, Joseph Y and Henao, Ricardo and Rubin, Geoffrey D and Carin, Lawrence},
  journal={Medical image analysis},
  volume={67},
  pages={101857},
  year={2021},
  publisher={Elsevier}
}

@inproceedings{molino2025any,
  title={{Any-to-Any Vision-Language Model for Multimodal X-ray Imaging and Radiological Report Generation}},
  author={Molino, Daniele and Di Feola, Francesco and Shen, Linlin and Soda, Paolo and Guarrasi, Valerio},
  booktitle={2025 International Joint Conference on Neural Networks (IJCNN)},
  pages={1--8},
  year={2025},
  organization={IEEE}
}

@article{tiu2022expert,
  title={{Expert-level detection of pathologies from unannotated chest X-ray images via self-supervised learning}},
  author={Tiu, Ekin and Talius, Ellie and Patel, Pujan and Langlotz, Curtis P and Ng, Andrew Y and Rajpurkar, Pranav},
  journal={Nature biomedical engineering},
  volume={6},
  number={12},
  pages={1399--1406},
  year={2022},
  publisher={Nature Publishing Group UK London}
}

@article{chen2022fast,
  title={{Fast and scalable search of whole-slide images via self-supervised deep learning}},
  author={Chen, Chengkuan and Lu, Ming Y and Williamson, Drew FK and Chen, Tiffany Y and Schaumberg, Andrew J and Mahmood, Faisal},
  journal={Nature Biomedical Engineering},
  volume={6},
  number={12},
  pages={1420--1434},
  year={2022},
  publisher={Nature Publishing Group UK London}
}

@inproceedings{zhang2012query,
  title={{Query specific fusion for image retrieval}},
  author={Zhang, Shaoting and Yang, Ming and Cour, Timothee and Yu, Kai and Metaxas, Dimitris N},
  booktitle={Computer Vision--ECCV 2012: 12th European Conference on Computer Vision, Florence, Italy, October 7-13, 2012, Proceedings, Part II 12},
  pages={660--673},
  year={2012},
  organization={Springer}
}

@article{saxena2021generative,
  title={{Generative adversarial networks (GANs) challenges, solutions, and future directions}},
  author={Saxena, Divya and Cao, Jiannong},
  journal={ACM Computing Surveys (CSUR)},
  volume={54},
  number={3},
  pages={1--42},
  year={2021},
  publisher={ACM New York, NY, USA}
}

@article{amirrajab2025radiology,
  title={{Radiology Report Conditional 3D CT Generation with Multi Encoder Latent diffusion Model}},
  author={Amirrajab, Sina and Salahuddin, Zohaib and Kuang, Sheng and Woodruff, Henry C and Lambin, Philippe},
  journal={arXiv preprint arXiv:2509.14780},
  year={2025}
}

@article{huang2019clinicalbert,
  title={{Clinicalbert: Modeling clinical notes and predicting hospital readmission}},
  author={Huang, Kexin and Altosaar, Jaan and Ranganath, Rajesh},
  journal={arXiv preprint arXiv:1904.05342},
  year={2019}
}

@misc{balachandran2024medembed,
  author = {Balachandran, Abhinand},
  title = {{MedEmbed: Medical-Focused Embedding Models}},
  year = {2024},
  url = {https://github.com/abhinand5/MedEmbed}
}

@article{zhang2023biomedclip,
  title={{Biomedclip: a multimodal biomedical foundation model pretrained from fifteen million scientific image-text pairs}},
  author={Zhang, Sheng and Xu, Yanbo and Usuyama, Naoto and Xu, Hanwen and Bagga, Jaspreet and Tinn, Robert and Preston, Sam and Rao, Rajesh and Wei, Mu and Valluri, Naveen and others},
  journal={arXiv preprint arXiv:2303.00915},
  year={2023}
}

@misc{challenge,
  author       = {Hamamci, Ibrahim Ethem and
                others},
  title        = {{Challenge for Vision-Language Modeling in 3D
                   Medical Imaging (VLM3D)
                  }},
  month        = mar,
  year         = 2025,
  publisher    = {Zenodo},
  doi          = {10.5281/zenodo.15052708},
}

@article{robinson2020contrastive,
  title={{Contrastive learning with hard negative samples}},
  author={Robinson, Joshua and Chuang, Ching-Yao and Sra, Suvrit and Jegelka, Stefanie},
  journal={arXiv preprint arXiv:2010.04592},
  year={2020}
}

@article{kalantidis2020hard,
  title={{Hard negative mixing for contrastive learning}},
  author={Kalantidis, Yannis and Sariyildiz, Mert Bulent and Pion, Noe and Weinzaepfel, Philippe and Larlus, Diane},
  journal={Advances in neural information processing systems},
  volume={33},
  pages={21798--21809},
  year={2020}
}

@article{hamamci2025better,
  title={{Better Tokens for Better 3D: Advancing Vision-Language Modeling in 3D Medical Imaging}},
  author={Hamamci, Ibrahim Ethem and Er, Sezgin and Shit, Suprosanna and Reynaud, Hadrien and Yang, Dong and Guo, Pengfei and Edgar, Marc and Xu, Daguang and Kainz, Bernhard and Menze, Bjoern},
  journal={arXiv preprint arXiv:2510.20639},
  year={2025}
}

@article{blankemeier2024merlin,
  title={{Merlin: A vision language foundation model for 3d computed tomography}},
  author={Blankemeier, Louis and Cohen, Joseph Paul and Kumar, Ashwin and Van Veen, Dave and Gardezi, Syed Jamal Safdar and Paschali, Magdalini and Chen, Zhihong and Delbrouck, Jean-Benoit and Reis, Eduardo and Truyts, Cesar and others},
  journal={Research Square},
  pages={rs--3},
  year={2024}
}

@inproceedings{kamath2023text,
  title={{Text encoders bottleneck compositionality in contrastive vision-language models}},
  author={Kamath, Amita and Hessel, Jack and Chang, Kai-Wei},
  booktitle={Proceedings of the 2023 Conference on Empirical Methods in Natural Language Processing},
  pages={4933--4944},
  year={2023}
}

@article{mei2022radimagenet,
  title={{RadImageNet: an open radiologic deep learning research dataset for effective transfer learning}},
  author={Mei, Xueyan and Liu, Zelong and Robson, Philip M and Marinelli, Brett and Huang, Mingqian and Doshi, Amish and Jacobi, Adam and Cao, Chendi and Link, Katherine E and Yang, Thomas and others},
  journal={Radiology: Artificial Intelligence},
  volume={4},
  number={5},
  pages={e210315},
  year={2022},
  publisher={Radiological Society of North America}
}

@article{chen2019med3d,
  title={{Med3d: Transfer learning for 3d medical image analysis}},
  author={Chen, Sihong and Ma, Kai and Zheng, Yefeng},
  journal={arXiv preprint arXiv:1904.00625},
  year={2019}
}

@article{song2021denoising,
  title={{Denoising diffusion implicit models}},
  author={Song, Jiaming and Meng, Chenlin and Ermon, Stefano},
  journal={arXiv preprint arXiv:2010.02502},
  year={2020}
}
\end{document}